%% file: arxiv.tex
\documentclass[conference]{ieeetran}

  \usepackage{amsmath,amssymb,amsfonts,amsthm,mathrsfs}

  \newcommand{\R}{\mathbb{R}}

  \providecommand{\norm}[1]{\|#1\|}

  \newcommand{\N}{\mathrm{N}}

  \usepackage{bm}
  
  \newcommand{\mbf}[1]{\mathbf{#1}}
  \newcommand{\vect}[1]{\mbf{#1}}
  \newcommand{\vectb}[1]{\bm{#1}}

  \newcommand{\eg}{\textit{e.g.}}
  \newcommand{\ie}{\textit{i.e.}}
  
  \newcommand{\etc}{\textit{etc.}}

  \usepackage{xcolor}

  \usepackage{graphicx}
  \graphicspath{ {./}{./fig/} }

  \usepackage{tikz,pgfplots}
  \usetikzlibrary{plotmarks,shapes,arrows}
  \pgfplotsset{compat=newest}

  \newlength\figureheight
  \newlength\figurewidth

  \usepackage{hyperref}
  \hypersetup{
    bookmarks=true,
    pdfstartview={FitH},
    colorlinks=true,
    linkcolor=black,
    citecolor=black,
    filecolor=black,
    urlcolor=black
  }

  \usepackage{color}

  \usetikzlibrary{external}

  \makeatletter
  \let\NAT@parse\undefined
  \makeatother

  \makeatletter
  \let\NAT@parse\undefined
  \makeatother
  \usepackage[square,numbers,sort&compress]{natbib}

\title{Inertial Odometry on Handheld Smartphones}

\author{
  \IEEEauthorblockN{Arno Solin}
  \IEEEauthorblockA{Aalto University\\
  Espoo, Finland\\
  arno.solin@aalto.fi}
  \and 
  \IEEEauthorblockN{Santiago Cortes}
  \IEEEauthorblockA{Aalto University\\
  Espoo, Finland\\
  santiago.cortesreina@aalto.fi}
  \and
  \IEEEauthorblockN{Esa Rahtu}
  \IEEEauthorblockA{Tampere Univ.\ of Tech.\\
  Tampere, Finland\\
  esa.rahtu@tut.fi}
  \and
  \IEEEauthorblockN{Juho Kannala}
  \IEEEauthorblockA{Aalto University\\
  Espoo, Finland\\
  juho.kannala@aalto.fi}
}

\begin{document}

\maketitle
\thispagestyle{empty}
\pagestyle{empty}

\begin{abstract}
  Building a complete inertial navigation system using the limited quality data provided by current smartphones has been regarded challenging, if not impossible. This paper shows that by careful crafting and accounting for the weak information in the sensor samples, smartphones are capable of pure inertial navigation. We present a probabilistic approach for orientation and use-case free inertial odometry, which is based on double-integrating rotated accelerations. The strength of the model is in learning additive and multiplicative IMU biases online. We are able to track the phone position, velocity, and pose in real-time and in a computationally lightweight fashion by solving the inference with an extended Kalman filter. The information fusion is completed with zero-velocity updates (if the phone remains stationary), altitude correction from barometric pressure readings (if available), and pseudo-updates constraining the momentary speed. We demonstrate our approach using an iPad and iPhone in several indoor dead-reckoning applications and in a measurement tool setup.
\end{abstract}

\section{Introduction}
\label{sec:intro}
\noindent
The deployment of global navigation satellite systems (GNSSs) has solved many large-scale positioning problems. However, these systems are not suited for precise tracking or for indoor use, which is where people spend most of their time. Accurate and fast indoor localization and tracking has many potential uses, including safety and emergency assistance, security, resource efficiency, navigation and augmented reality.

The idea of an inertial navigation system (INS, see \cite{Jekeli:2001,Britting:2010}) is to use the fusion of inertial sensors (accelerometers and gyroscopes) to continuously estimate the position, orientation, and velocity of a moving object. This type of tracking, known as dead-reckoning, is typically associated with aircraft, submarines, and missile technology. Recent advances in MEMS sensors have brought motion and rotation sensors to standard consumer smartphones and devices, and introduced the potential for new INS applications.

Smartphones and tablet devices are equipped with MEMS sensors in order to enhance human-computer interaction and enable new applications. For example, thanks to the accelerometer, devices can automatically rotate the screen based on the device orientation with respect to gravity. Furthermore, gyroscopes have enabled new ways to interact with digital content, such as watching of panoramic video or controlling games by rotating the device. In fact, besides gravitation sensing and tracking \cite{Sarkka+Tolvanen+Kannala+Rahtu:2015}, information fusion from accelerometers, gyroscopes and magnetometers can be utilised for robust real-time tracking of the full device orientation \cite{Madgwick+Harrison+Vaidyanathan:2011, Renaudin+Combettes:2014}. Such approaches are sometimes referred to as \emph{attitude and heading reference systems} (AHRS).

\begin{figure*}[!t]
  \tikzexternaldisable

  \tikzsetnextfilename{tikz-intro}

  \setlength{\figurewidth}{.65\textwidth}
  \setlength{\figureheight}{0.4780\figurewidth}

  \pgfplotsset{
    trim axis right,
    yticklabel style={rotate=90},
  }
  \footnotesize\centering%
  \hspace*{\fill}
  \input{./fig/intro.tex}
  \hspace*{\fill}
  \tikzsetnextfilename{tikz-person}%
  \begin{tikzpicture}
    \node[anchor=south west,inner sep=0] (image) at (0,0) %
      {\includegraphics[width=.4\columnwidth,keepaspectratio]{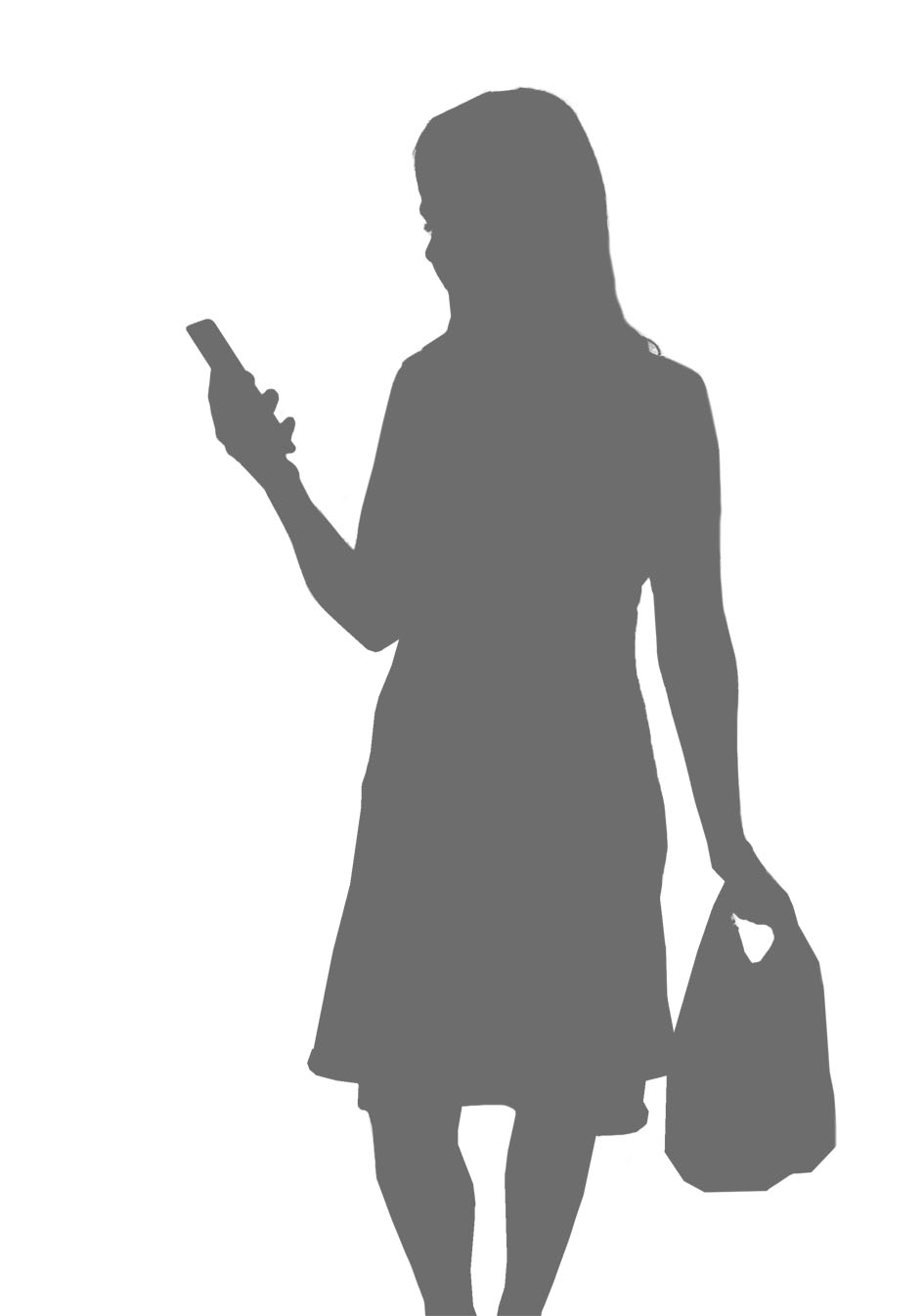}};
    \begin{scope}[x={(image.south east)},y={(image.north west)}]

      \tikzstyle{label} = [text width=1.5cm, align=center]
      \tikzstyle{line}  = [draw, very thick, blue!50]
      \tikzstyle{point} = [circle,draw,very thick,blue!50,fill=none,
                           minimum size=5mm,inner sep=0]

      \node [label] (lab_hand)   at (0.05,0.90) {Phone in hand};
      \node [label] (lab_pocket) at (0.05,0.10) {Phone in pocket};
      \node [label] (lab_bag)    at (0.95,0.45) {Phone in bag};

      \node [point] (hand)       at (0.25,0.70) {};
      \node [point] (pocket)     at (0.45,0.45) {};
      \node [point] (bag)        at (0.80,0.15) {};

      \path [line] (lab_hand)   |- (hand);
      \path [line] (lab_pocket) |- (pocket);
      \path [line] (lab_bag)    |- (bag);

    \end{scope}
  \end{tikzpicture}
  \hspace*{\fill}
  \vspace{-1em}
  \tikzexternalenable
  \caption{Features of the INS system summarized into one figure: The path was started on the ground floor with zero-velocity updates. After walking up the stairs to the first floor a position fix was given, after which the phone was put in a \textbf{closed bag}. Then the phone was put in the \textbf{pocket}. Before descending to the ground floor, the phone was taken out of the pocket and a second position fix was given (aligning the path to the map). On the ground floor a manual loop-closure was given. The data was collected by an iPhone~6 and calibrations were performed on the fly.}
  \label{fig:intro} 
\end{figure*}

Tracking the translational motion of devices based on inertial sensors is considerably harder than orientation tracking. However, certain applications, like pedestrian tracking and indoor positioning, would greatly benefit from accurate inertial navigation on smartphones. The difficulty of inertial navigation is due to the need to double-integrate the observed accelerations, which rapidly accumulates errors from the high noise-level of MEMS accelerometers. Small errors in the attitude estimation will make this even more challenging as the gravitation may `leak' to the integrated accelerations~\cite{Sachs:2010}. 

In order to solve the aforementioned challenges, many current systems resort to additional hardware, such as foot-mounted sensors \cite{Foxlin:2005,Nilsson+Zachariah+Skog+Handel:2013} or video cameras.
While providing accurate results, these are quite impractical for wide use in consumer applications. For example, camera-based approaches do not work when the device is in a closed bag or pocket, and capturing and processing video consumes a lot of energy compromising battery longevity. Further, while foot-mounted sensors can provide accurate tracking thanks to frequent zero-velocity updates and high-quality sensors, they are inconvenient for large-scale consumer use and the current solutions do not work well when the movement happens without steps, for example in a trolley, elevator, or escalator.

In this paper we show how an inertial navigation system can be built to work on the limited-quality data provided by a standard smartphone. We propose a general inertial navigation approach which is not based on detecting steps and therefore works in various use cases, covering both legged motion and motion with wheels, as well as motion in elevators and escalators. Moreover, the approach does not require constraining the device orientation, and thus the device can be held freely. In addition, the approach is computationally light-weight and capable for real-time processing on a smartphone. To the best of our knowledge, this is the first paper demonstrating such a system with a standard smartphone.

Figure~\ref{fig:intro} summarizes the features of the proposed INS system in a test performed with a standard iPhone~6. In this example, the path was started on the ground floor with zero-velocity updates for calibrating the sensors (no pre-calibrations done). After walking up the stairs to the first floor holding the phone in the hand, a position fix was given, after which the phone was put in a bag. Next, the phone was taken out of the bag and put in the trouser pocket. Before descending to the ground floor, the phone was taken out of the pocket and a second position fix was given, which aligned the path to the map. On the ground floor a manual loop-closure indicated that we were where we started.

The contributions of this paper are two-fold: 
\begin{itemize}
  \item We show that inertial navigation on a standard smartphone is feasible by careful crafting of the model, taking advantage of weak signals, and accounting for uncertainties in data.
  \item We present a streamlined estimation approach for the INS problem which builds upon learning the dynamical sensor bias parameters as a part of the state variables. The probabilistic inference is solved by a sequential filtering scheme, where the only approximations come from the linearizations inside the extended Kalman filter. The approach is complemented with zero-velocity updates and pseudo-measurements limiting the momentary speed. 
\end{itemize} 

This paper is structured as follows. In the next section we provide a brief literature review of previous work. In Section~\ref{sec:methods} we present the INS model. The exact model is presented in detail, and measurement updates for fusing measurements with dynamics are described. Section~\ref{sec:experiments} presents empirical studies where the inertial navigation algorithm is employed in pedestrian dead-reckoning examples, a generalized dead-reckoning example, and as a measurement tool.  Finally, the results are discussed in Section~\ref{sec:discussion}.

\section{Related Work}
\label{sec:literature}
\noindent
Inertial navigation systems have been studied for decades. The classical literature cover primarily navigation applications for aircraft and large vehicles \cite{Jekeli:2001,Bar-Shalom+Li+Kirubarajan:2001,Titterton+Weston:2004,Britting:2010}. The development of handheld consumer-grade devices has awakened an interest in pedestrian navigation applications, where the challenges are slightly different from those in the classical approaches. That is, the limited quality of smartphone MEMS sensors and abrupt motions of hand-held devices pose additional challenges which have so far prevented generic inertial navigation solutions for smartphone applications.

In order to focus on the relevant previous literature, we restrict our scope to tracking algorithms that use the sensors available in a smartphone, primarily accelerometers, gyroscopes, and magnetometers.

The extensive survey by Harle \cite{Harle:2013} covers many approaches with different constraints for the use of inertial sensors for pedestrian dead-reckoning (PDR). Typically INS systems either constrain the motion model or rely on external sensors. In fact, we are not aware of any previous system which would have all the capabilities that we demonstrate in this paper.

One prominent INS solution relying on external hardware is the OpenShoe project \cite{Nilsson+Zachariah+Skog+Handel:2013, Nilsson+Gupta+Handel:2014}. It uses foot-mounted inertial sensors with several pairs of accelerometers and gyroscopes to estimate the step-by-step PDR (in an INS-SHS framework, see below). The model is constrained by zero-velocity updates (ZUPTs) on each step once the foot touches the ground.

Step and heading systems (SHS, see \eg\ \cite{Woodman:2010,Renaudin+Combettes:2014, Kang+Han:2015, Yuan+Yu+Zhang+Wang+Liu:2015, Chen+Meng+Wang+Zhang+Tian+Yang:2015}) use the inertial sensor to estimate the heading and the step length of the user. These are introduced into a constrained model that estimates the walking path by accumulating the step vectors in order to do PDR. These systems have been proven to work well for PDR in short and medium range but they typically impose constraints for the device orientation. For example, the device orientation is often known or fixed with respect to the walking direction. Further, they are very sensitive to changing gaits and are prone to false positives (see discussion in \cite{Harle:2013}). A recent approach \cite{Xiao+Wen+Markham+Trigoni:2014} uses bipedal locomotion models to model the periodical behavior of the INS in a smartphone and, thus, estimate steps. Although they are able to relax the constraint of known and fixed device orientation to some extent, their approach is still step-based, and heading estimation is error-prone, especially if there are frequent and abrupt changes in orientation.

Besides inertial PDR systems, there exist many camera-aided inertial tracking solutions (visual-inertial odometry), which can provide accurate tracking in visually distinguishable environments (\eg\ \cite{Li+Kim+Mourikis:2013, Hesch+Kottas+Bowman+Roumeliotis:2014, Bloesch+Omari+Hutter+Siegwart:2015,Solin+Cortes+Rahtu+Kannala:2018-WACV}). However, as these approaches require constant use of a video camera, causing increased battery usage, and unobstructed visibility of surroundings, they are not directly comparable to our approach.

Finally, it should be noted that often odometry estimation techniques, either inertial or visual, are part of larger localization systems, which combine odometry with various kinds of maps or fingerprinting methods that provide reference positions. Examples of mapped signals, which have been utilized for indoor localization, include signal strengths of Wi-Fi and Bluetooth radio beacons \cite{Mirowski+Ho+Yi+MacDonald:2013}, cellular communications radio \cite{Martin+Vinyals+Friedland+Bajcsy:2010}, RFID tags \cite{Ruiz+Granja+Honorato+Rosas:2012}, and variations of the ambient magnetic field \cite{Solin+Kok+Wahlstrom+Schon+Sarkka:2015,Solin+Sarkka+Kannala+Rahtu:2016}.

\section{Methods}
\label{sec:methods}

\noindent

Even though, the physical interpretation of how an inertial navigation system works is straight-forward, this setup has many pitfalls. All inertial navigation systems suffer from integration drift. Small errors in the measurements of acceleration and angular velocity cause progressively larger errors in velocity---and even greater errors in position. The dominating component in the accelerometer data is gravity, which means that even slight errors in orientation make the gravity `leak' into the estimates. The sequential nature of the problem makes the errors accumulate. Once the estimates start to drift, they quickly diverge.

These problems underline the importance of accurately modelling and handling the inherent noises, sampling times, uncertainties, and numerical instabilities in the system. We use the data provided by the inertial measurement unit (IMU) in the smartphone to continuously infer the relative change in position, velocity and orientation of the device with respect to a starting point (see \cite{Jekeli:2001,Britting:2010}). The three-axis IMU measures data of the specific force (accelerometer data) and angular rate (gyroscope data).

\subsection{Non-Linear Estimation}
\label{sec:nonlin-estimation}
\noindent
An inertial navigation system is non-linear both in the dynamics and observations. Non-linear filtering methods (see \cite{Sarkka:2013} for an overview) are concerned with this kind of estimation problems. Consider a non-linear state-space equation model of form
\begin{align}
  \vect{x}_k &= \vect{f}_k(\vect{x}_{k-1}, \vectb{\varepsilon}_k), \label{eq:dynamic} \\
  \vect{y}_k &= \vect{h}_k(\vect{x}_k, \vectb{\gamma}_k), \label{eq:measurement} 
\end{align}
where $\vect{x}_k \in \R^n$ is the state at time step $t_k$, $k=1,2,\ldots$, $\vect{y}_k \in \R^m$ is a measurement, $\vectb{\varepsilon}_k \sim \N(\vectb{0}, \vect{Q}_k)$ is the Gaussian process noise, and  $\vectb{\gamma}_k \sim \N(\vectb{0},\vect{R}_k)$ is the Gaussian measurement noise. The dynamics and measurements are specified in terms of the dynamical model function $\vect{f}_k(\cdot)$ and the measurement model function $\vect{h}_k(\cdot)$, both of which can depend on the time step $k$.

We employ the extended Kalman filter (EKF, \cite{Bar-Shalom+Li+Kirubarajan:2001}) which provides a means of approximating the state distributions 
\begin{equation}
  p(\vect{x}_k \mid \vect{y}_{1:k}) \simeq \N(\vect{x}_k \mid \vect{m}_{k}, \vect{P}_{k})
\end{equation}
with Gaussians through first-order linearizations. In the experiments, we also employ the fixed-interval extended Rauch--Tung--Striebel smoother (see \cite{Sarkka:2013} for detailed presentation) for obtaining the state distributions $p(\vect{x}_k \mid \vect{y}_{1:N})$ conditioned on the entire track of observations.

\subsection{Dynamical Model}
\noindent
The state variables hold the knowledge of the system state at any given time step. The state variables are:
\begin{equation}
  \vect{x}_k = (\vect{p}_k, \vect{v}_k, \vect{q}_k, \vect{b}_k^\mathrm{a}, \vect{b}_k^{\omega}, \vect{T}_k^\mathrm{a}),
\end{equation}
where $\vect{p}_k \in \R^3$ is the position, $\vect{v}_k \in \R^3$ the velocity, and $\vect{q}_k$ the orientation unit quaternion at time step $t_k$. The remaining components are the additive accelerometer and gyroscope bias components, and $\vect{T}_k^\mathrm{a}$ denotes the diagonal multiplicative scale error of the accelerometer.

The dynamical model (Eq.~\ref{eq:dynamic}) is based on the assumption that position is velocity once integrated, and velocity is acceleration (with the influence of gravity removed) once integrated. The orientation of the acceleration is tracked with gyroscope measurements. The accelerometer and gyroscope readings are regarded as control signals, and their measurement noises are seen as the process noise of the system.

The dynamical model given by the mechanization equations (see, \eg, \cite{Titterton+Weston:2004,Nilsson+Zachariah+Skog+Handel:2013} for similar model formulations) is
\begin{equation}\label{eq:ins-model}
  \begin{pmatrix}
    \vect{p}_k \\ \vect{v}_k \\ \vect{q}_k
  \end{pmatrix}
  =
  \begin{pmatrix}
    \vect{p}_{k-1} + \vect{v}_{k-1}\Delta t_k \\
    \vect{v}_{k-1} + [\vect{q}_k (\tilde{\vect{a}}_k + \vectb{\varepsilon}^\mathrm{a}_k) \vect{q}_k^\star - \vect{g}] \Delta t_k \\
    \vectb{\Omega}[(\tilde{\vectb{\omega}}_k + \vectb{\varepsilon}^\omega_k) \Delta t_k] \vect{q}_{k-1}
  \end{pmatrix},
\end{equation}
where the time step length is given by $\Delta t_k = t_{k} - t_{k-1}$ (note that we {\em do not} assume equidistant sampling times), the accelerometer input is denoted by $\tilde{\vect{a}}_k$ and the gyroscope input by $\tilde{\vectb{\omega}}_k$. Gravity $\vect{g}$ is a constant vector. The quaternion rotation is denoted by the $\vect{q}_k [\cdot] \vect{q}_k^\star$ notation, and the quaternion rotation update is given by the function $\vectb{\Omega}: \R^3 \to \R^{4 \times 4}$ (see \cite{Titterton+Weston:2004} for details).

The system is deterministic up to the uncertainties (measurement noises and biases) associated with the accelerometer and gyroscope data. The process noises associated with the inputs are modelled as i.i.d.\ Gaussian noise $\vectb{\varepsilon}^\mathrm{a}_k \sim \N(\vectb{0},\vectb{\Sigma}^\mathrm{a} \Delta t_k)$ and $\vectb{\varepsilon}^\omega_k \sim \N(\vectb{0},\vectb{\Sigma}^\omega \Delta t_k)$. The Jacobians of \eqref{eq:ins-model}, required for the linearizations in filtering, can be constructed in closed-form.

The accelerometer and gyroscope readings provided by the low-cost sensors in the mobile device may suffer from misalignment errors and scale errors in addition to white measurement noise. These are taken into account inside the dynamic model as follows:
\begin{equation}
\begin{split}
  \tilde{\vect{a}}_k &= \vect{T}_k^\mathrm{a} \, \vect{a}_k - \vect{b}^\mathrm{a}_k, \\
  \tilde{\vectb{\omega}}_k &= \hphantom{\vect{T}_k^\mathrm{a} \, }\vectb{\omega}_k - \vect{b}^\omega_k,
\end{split}
\end{equation}
where the accelerometer and gyroscope sensor readings at $t_k$ are $\vect{a}_k$ and $\vectb{\omega}_k$. The additive biases are denoted by $\vect{b}^\mathrm{a}_k$ and $\vect{b}^\omega_k$, respectively. The diagonal scale error matrix $\vect{T}_k^\mathrm{a}$ accounts for miscalibrations in the accelerometer scale.

The biases and diagonal scale error terms are estimated online as a part of the state estimation problem. They are considered fixed over the entire time horizon, thus the dynamic model for their part is fixed and without any process noise: 
\begin{equation}
  \vect{b}^\mathrm{a}_k = \vect{b}^\mathrm{a}_{k-1}, \quad
  \vect{b}^{\omega}_k = \vect{b}^{\omega}_{k-1}, \quad \text{and} \quad
  \vect{T}_k^\mathrm{a} = \vect{T}_{k-1}^\mathrm{a}.
\end{equation}
This means that their values are controlled by the prior state and information provided by the measurement updates.

The complete dynamical model must be differentiated both in terms of the state variables and process noise terms in order to fit the EKF estimation scheme (see  \cite{Sarkka:2013}). These derivatives can be derived in closed-form in order to preserve the stability of the system.
The initial (prior) state is given by $\vect{p}_0 \sim \N(\vect{0}, \vectb{\Sigma}_0^\mathrm{p})$,  $\vect{v}_0 \sim \N(\vect{0}, \vectb{\Sigma}_0^\mathrm{v})$, and  $\vect{q}_0$ chosen such that it defines the initial orientation (deduced from gravity direction). The additive biases are initialized to zero and the scale bias to an identity matrix.

\begin{figure*}[!t]
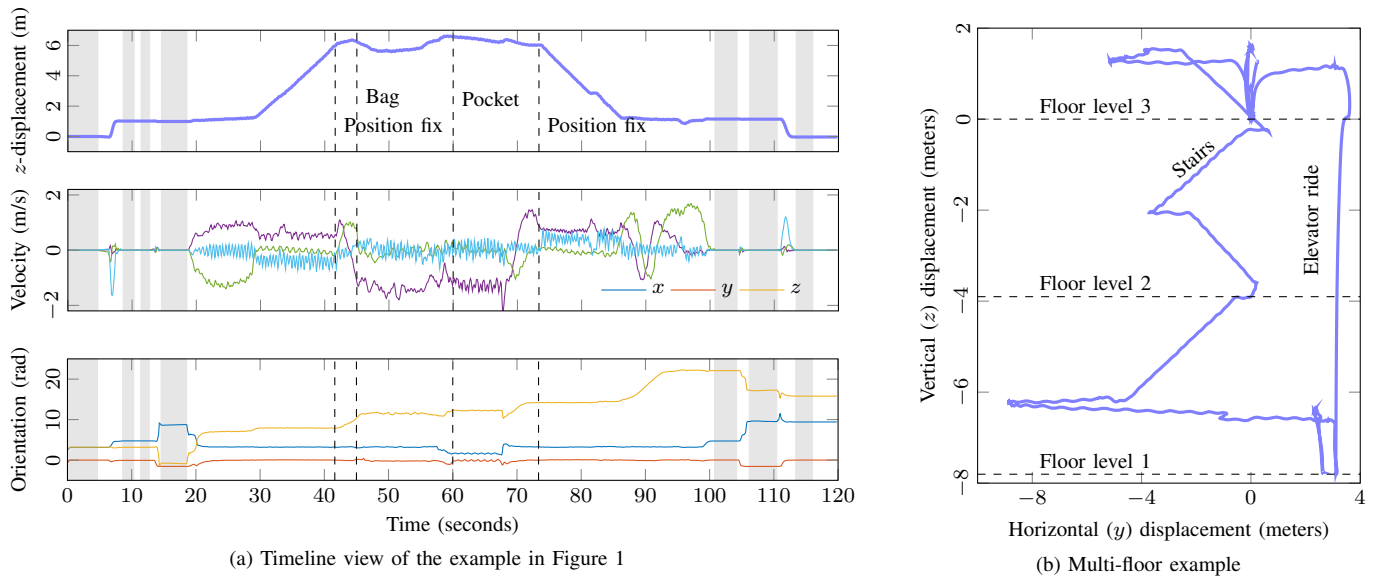

  \begin{minipage}{.66\textwidth}

  \setlength{\figurewidth}{.90\textwidth}
  \setlength{\figureheight}{0.15\figurewidth}

  \pgfplotsset{
    trim axis right,
    yticklabel style={rotate=90},
    legend columns=3,
    legend style={draw=none},
    ylabel absolute,
    ylabel style={yshift=-0.6cm}
  }

  \footnotesize
  \tikzsetnextfilename{tikz-altitude}%
  \input{./fig/altitude.tex} \\ 
  \tikzsetnextfilename{tikz-velocity}%
  \input{./fig/velocity.tex} \\
  \tikzsetnextfilename{tikz-orientation}%
  \input{./fig/orientation.tex}
  \vspace*{1em}
  \hspace{3cm}{\footnotesize {(a)~Timeline view of the example in Figure~1}} \\
  \vspace*{0.1em}
  \end{minipage}%
  \hspace*{0em}
  \begin{minipage}{.33\textwidth}

  \tikzsetnextfilename{tikz-vertical}

  \setlength{\figurewidth}{.85\columnwidth}
  \setlength{\figureheight}{1.19\figurewidth}

  \pgfplotsset{
    trim axis right,
    yticklabel style={rotate=90},
  }

  \footnotesize\centering%
  \input{./fig/vertical.tex}
  \vspace*{1em}
  {\footnotesize {(b)~Multi-floor example}}\\

  \end{minipage}%
  \tikzexternalenable
  \vspace*{-1.4em}
  \caption{(a)~The altitude (vertical) profile, the velocities, and the orientations of the phone along the path in Figure~\ref{fig:intro}. The shading shows the stationarity detection outcome, where zero-velocity updates were triggered. The subtle periodicity in the path is due to walking, and the drop in altitude on the first floor is because of the phone being in a bag for parts of the path. (b)~A PDR example with first descending two levels down and then taking the elevator back up. The path was started at origin and the path ends with a loop-closure in the same place. The points where the phone touches the floor level (the sharp drops in vertical position) are zero-velocity updates. No absolute position info was given to the model.}
  \label{fig:states} 
  \vspace*{-.5em}
\end{figure*}

\subsection{Position Fixes and Loop-Closures}
\noindent
In terms of the sequential inference scheme all auxiliary observation data is combined with the model through the measurement model in Equation~\ref{eq:measurement}. Position fixes are noisy measurements of the position vectors $\vect{p}_k$ in the state (\ie\ $\vect{h}_\mathrm{pos.}(\vect{x}) = \vect{p}$). The additive Gaussian measurement noise represents the uncertainty associated with the given position.

Position fixes provide uncertain information of the position and thus also the distance travelled between the position fixes. This first-hand information helps the model pin down the bias estimates very accurately. Loop-closure points do not provide any exact position information, but indicate that at two different points in time, the positions are the same. Also this information is valuable in inferring sensor biases.

Manual loop-closures can be combined with the estimation scheme by augmenting the current position estimate in the state by a Kalman update at loop-opening. The state dimension grows by three at opening the loop, and the state becomes
\begin{equation}
  \vect{x} = (\vect{x}^\mathrm{old}, \vect{p}^\mathrm{LC}), 
\end{equation}
where the prior $\vect{p}^\mathrm{LC} \sim \N(\vect{0}, \vectb{\Sigma}_0^\mathrm{LC})$ with the $ \vectb{\Sigma}_0^\mathrm{LC}$ sufficiently large indicating the non-informativity of the initial location of the loop-closure point. In practice, both at $t_\mathrm{open}$ and $t_\mathrm{close}$ (the loop can be closed many times) the measurement model
\begin{equation}
  \vect{h}_\mathrm{LC}(\vect{x}) = \vect{p} - \vect{p}^\mathrm{LC}
\end{equation}
defines an observation $\vect{y}=\vect{0}$ with some measurement noise $\vectb{\gamma} \sim \N(\vect{0}, \vectb{\Sigma}^\mathrm{LC})$. The measurement noise covariance $\vectb{\Sigma}^\mathrm{LC}$ should reflect the mismatch of the user not exactly being at the loop-closure spot. Both the position fix and loop-closures are linear observations of the state, and can thus be implemented using a standard (linear) Kalman update.

\subsection{Zero-Velocity Updates}
\noindent
In this paper, the most important source of auxiliary information is so called zero-velocity updates (ZUPTs, see \cite{Nilsson+Zachariah+Skog+Handel:2013} for discussion). Once the phone is detected to be stationary for any period of time, it is known to the model that the system velocity must be zero ($\vect{v}=\vect{0}$). In terms of the measurement model, this means
\begin{equation}
  \vect{h}_\mathrm{ZUPT}(\vect{x}) = \vect{v}
\end{equation}
and the additive measurement noise $\vectb{\gamma}$ is small (by only specifying a pseudo-noise scale). This update can be performed as a standard (linear) Kalman update.

For triggering, we use an iterative Dickey--Fuller stationarity test \cite{Dickey+Fuller:1979} on a rolling window of accelerometer data (window size 250~ms) with an additional requirement of the sample standard deviation being small. This means that trends in the data are used as a proxy for movement.

\subsection{Pseudo-Measurement Updates}
\noindent
Without position fixes, loop-closures, or ZUPTs the inertial navigation system quickly becomes unstable. Once the estimates start diverging, they easily loose their numerical precision. The main source of these problems is gravity `leaking' into the acceleration input and corrupting the velocity vector. Once the velocity starts to drift, the position diverges almost instantly.
However, even without other auxiliary information, it is possible to keep the system informed about a reasonable scale of velocity. In our model, we present a simple yet powerful pseudo-update formulation that keeps the speed in the range of some meters per second and discourages the system from accelerating into higher velocities.

The pseudo-update model is defined in terms of the speed of the object, when it is not stationary. The speed (the Euclidean norm of the velocity) is
\begin{equation}
  h_\mathrm{pseudo}(\vect{x}) = \norm{\vect{v}}.
\end{equation}
In our experiments the pseudo-updates are parametrized as follows. The speed observation $y = 0.75$~m/s with a measurement noise $\gamma = \N(0,2^2)$. The large measurement noise variance keeps the update non-informative in comparison to other information sources.

\subsection{Barometer Readings}
\noindent
Barometric air pressure data (typically also available in high-end smartphones) can be mapped to heights through linearization of the barometric formula around sea level. Over short time periods the air pressure at a given altitude tends to stay constant. In this case the barometer readings relative to the starting point can provide absolute height updates (corresponding to position fixes as presented above).

The barometric pressure drifts over longer time horizons (in the order of tens of minutes), leading to accumulation of measurement errors. Another approach is to only use the relative pressure differences between two consecutive barometer observations mitigating drift issues. This alternative corresponds to opening an altitude loop-closure point on each barometer observation and closing them on the next.

\section{Experiments}
\label{sec:experiments}
\noindent
In the examples, the interest was put on Apple phones and tablets---mostly because of their uniform hardware and good software compatibility between devices. The device models used in the examples are the iPhone~6 and the iPad~Pro (12.9-inch model). Both these models are equipped with built-in IMUs (InvenSense MP67B) and a barometric sensor (Bosch Sensortec BMP280). In all experiments, the IMU sensor data and the associated timestamps were collected at 100~Hz, and the barometer data (when used) at approximately 0.75~Hz. The data was collected using an in-house developed data collection application, and the paths were reconstructed on the iPhone hardware off-line after the data acquisition.

\subsection{Pedestrian Dead-Reckoning}
\noindent
The most apparent use case for the presented model is to apply it to pedestrian dead-reckoning, where the mobile phone (iPhone~6) is carried by the user indoors. There exist a multitude of methods for dead-reckoning using data provided by mobile phones. Therefore the aim of this experiment is to show how this method differs from others by its generality.

Figures~\ref{fig:intro} and \ref{fig:states}(b) summarize features of the proposed INS system; the example includes the use of zero-velocity updates, position fixes, pseudo-measurements constraining the speed, and barometer observations. This experiment covers traditional navigation-like PDR use cases (walking with the phone in a fixed orientation), where SHS systems are often used, cross-floor tracking, where visual tracking methods are usually the method of choice, and bag/pocket use cases, which currently often require resorting to radio based positioning. The generality of our INS system can cover them all with only one method and no external hardware.

In the first example, the path was started on the ground floor with zero-velocity updates  (no pre-calibrations done). First the user walked up a flight of stairs to the first floor holding the phone in the hand. On the first floor a position fix was given, after which the phone was put in a bag. Next, the phone was taken out of the bag and put in the trouser pocket. Before descending to the ground floor, the phone was taken out of the pocket and a second uncertain position observation was given, which aligned the path to the map and was able to provide absolute information of the scale. On the ground floor a manual loop-closure was given to indicate that the phone had returned to the starting point, and the phone was placed on the floor for some final zero-velocity updates. The tracking path is accurate and follows the true path up to decimetres. 

Figure~\ref{fig:states}(a) shows the altitude profile of the path. The ZUPTs where the phone is placed on the floor are clearly showing, as well as the stair climbing. The drop in altitude on the first floor is due to the phone being in the bag for a part of the path. The figure also shows the estimated velocity. The periodicity is due to walking. This effect is less evident when the phone is in the bag, and at clearest when the phone is in the trouser pocket firmly attached to the body.

We briefly present a second PDR example which is shown in Figure~\ref{fig:states}(b). In this example the path was started at origin with zero-velocity updates in different phone orientations. After this the user walked two floors down. When waiting for the elevator on floor level~1, further ZUPTs were done. The path is completed with taking the elevator back to floor level~3 and closing the loop at the starting point. In this example no absolute position information was given. The scale comes entirely from the accelerometer data. In both examples, an backward smoother pass (see Sec.~\ref{sec:nonlin-estimation}) is run after every update, thus also correcting the past estimates.

These expeiments demostrate the unconventional nature of the proposed method; this odometry method delivers a combination of use cases, which cannot be delivered with other methods running on the same device. Visual methods fail in the bag/pocket, and SHS methods fail when the device orientation is not fixed or steps/motion cannot be observed.

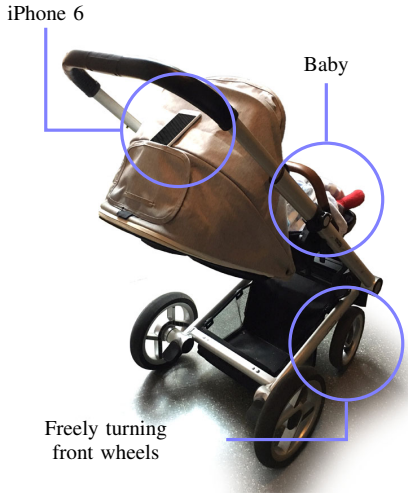
\begin{figure}[!t]   
  \centering\footnotesize

  \tikzsetnextfilename{tikz-stroller-setup}%
  \begin{tikzpicture}
    \node[anchor=south west,inner sep=0] (image) at (0,0) %
      {\includegraphics[width=.6\columnwidth,keepaspectratio]{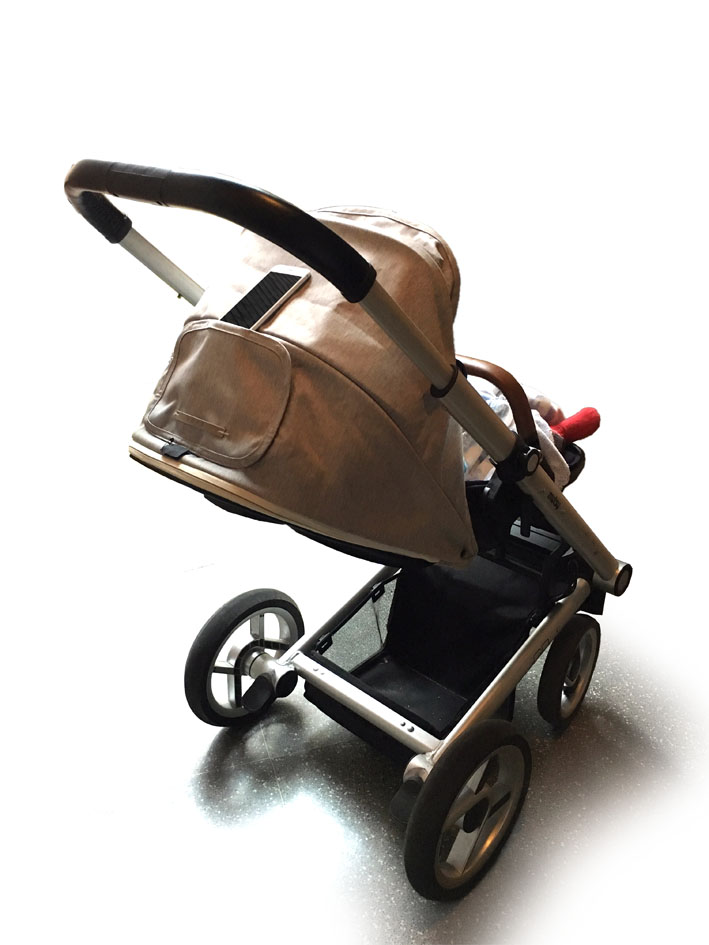}};
    \begin{scope}[x={(image.south east)},y={(image.north west)}]

      \tikzstyle{label} = [text width=3cm, align=center]
      \tikzstyle{line} = [draw, very thick, blue!50]
      \tikzstyle{point} = [circle,draw,very thick,blue!50,fill=none,minimum size=1.5cm,inner sep=0]

      \node [label] (lab_phone) at (0.05,0.90) {iPhone 6};
      \node [label] (lab_baby)  at (0.75,0.80) {Baby};
      \node [label] (lab_wheel) at (0.20,0.10) {Freely turning front wheels};

      \node [point] (phone)     at (0.38,0.68) {};
      \node [point] (baby)      at (0.75,0.55) {};
      \node [point] (wheel)     at (0.80,0.28) {};

      \path [line] (lab_phone) |- (phone);
      \path [line] (lab_baby) |- (baby);
      \path [line] (lab_wheel) -| (wheel);

    \end{scope}
  \end{tikzpicture}
  \tikzexternalenable

  \caption{The experiment setup with a wheeled baby pushchair/stroller and an iPhone placed on the top for tracking.}
  \label{fig:stroller-setup}
\end{figure} 

\begin{figure}[!t]
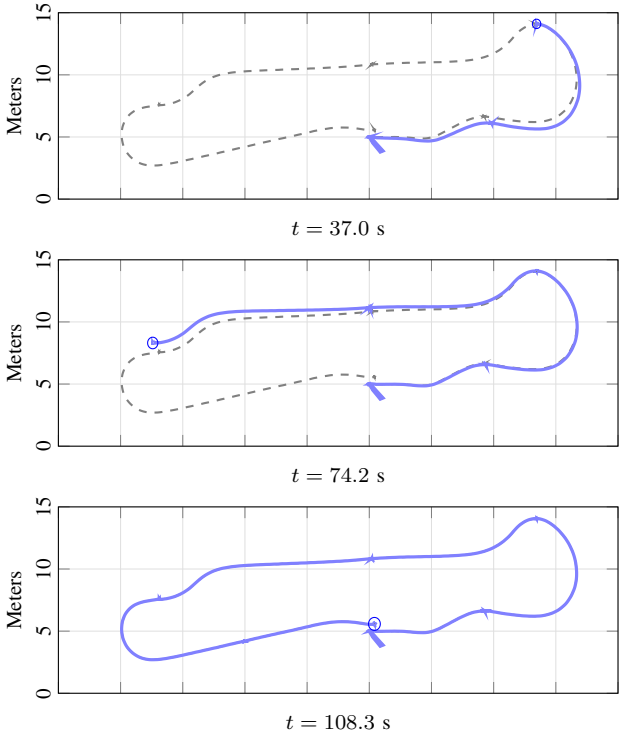

  \tikzexternaldisable

  \setlength{\figurewidth}{.84\columnwidth}
  \setlength{\figureheight}{0.7507\figurewidth}

  \pgfplotsset{
    trim axis right,
    yticklabel style={rotate=90},
    grid style={very thin,gray!25},
    ylabel={Meters}
  }

  \footnotesize\centering%

  \tikzsetnextfilename{tikz-stroller-2}%
  \input{./fig/stroller-2.tex}

  \tikzsetnextfilename{tikz-stroller-4}%
  \input{./fig/stroller-4.tex}

  \tikzsetnextfilename{tikz-stroller-6}%
  \input{./fig/stroller-6.tex}
  \tikzexternalenable
  \caption{The evolution of the position estimate and path are visualized along the way at those points where the pushchair was momentarily stopped. The dashed line shows the final path for reference. The phone remained leaning on the top for the entire experiment. No position fixes nor loop-closures were used.}
  \vspace*{-2em}
  \label{fig:stroller} 
\end{figure}

\subsection{Generalized Dead-Reckoning}
\noindent
Wheel based motion and general non-legged motion are use cases which are not covered by conventional step counting PDR methods. We now set the method in a more general scope for general dead-reckoning that can be applied to any wheeled, sliding, or flying objects indoors or outdoors. Applications include push-carts, trolleys, robots, hover boards, quadcopters, \etc\
We include an example with a human manoeuvred wheeled object with an intrinsic noise source---that is a baby pushchair/stroller with a baby on board. Figure~\ref{fig:stroller-setup} shows the test setup, where the phone is placed leaning on the top. The phone (iPhone~6) remains fixed to the pushchair body for the entire experiment.

Walking was started from a stationary state, where the phone automatically performed ZUPTs. Along the route the pushchair was stopped irregularly and ZUPTs triggered if it became stationary enough. INo position fixes nor loop-closures were used. The only measurement data are the automatic zero-velocity updates, the barometer observations, and pseudo-updates constraining the momentary speed. The total path length was {$\sim$}93~m.
Figure~\ref{fig:stroller} shows the path estimate at the times when the pushchair was stopped. The dashed line is the path estimate at $t=108.3$~s which is shown for reference. The zero-velocity updates in the various heading angles of the pushchair are clearly enough to capture the bias estimates and stabilize the system. The final estimate is off from the starting point by {$\sim$}0.73~m (0.78\%). 
As the phone orientation is fixed to the pushchair and only moves in a plane, this use case is well suited for SHS sytems \cite{Harle:2013}. For comparison we implemented a 2D odometry method combining movement detection with turn rates projected to the horizontal plane \cite{Sarkka+Tolvanen+Kannala+Rahtu:2015}. The final estimate is only 1.80~m (1.94\%) off from the starting point at the end, which is good for an SHS method.

\subsection{Comments on Computational Complexity}
\noindent
The odometry method was implemented in C++ with wrappers in Objective-C for running on the device. The implementation uses the Eigen matrix library. The computational complexity scales linearly with the number of data points, meaning a constant computational burden per sample.

For development purposes, the method was run on the device hardware, but not online. For example, running the odometry for the track in Figure~\ref{fig:stroller} (108.3~s of data) took 0.30~s on the iPhone~6 hardware (single-threaded). Thus the method is capable of running in a real-time application.

\subsection{Sensing of Surroundings}
\noindent
The model proposed in this paper has many potential applications beyond simple odometry and tracking. For example, the orientation and ZUPT information can be used as a measurement tool per se. By consecutively placing the phone flat on the walls of a room, a model of the geometry of the room can be built. From each ZUPT on the trajectory, a wall can be projected parallel to the phone screen, thus capturing the geometry of the room. Associating several ZUPTs to the same wall with the additional knowledge that the points span a plane through the space, can also make it possible to better estimate the wall placement and orientation.

The model is flexible enough that new constraints---in form of estimated quantities and prior information about them---can be introduced. For this particular application there are several useful constraints, such as coplanarity between some ZUPT positions. A similar smartphone application that delivers these functionalities is publicly available. Therefore we seek to deliver comparable results to the RoomScan (Locometric~Ltd, \url{http://locometric.com}) application.

Conventional loop-closures are not suited for this particular purpose. In this case the loop-closure points are touching the same plane. This plane is a line in the $xy$-plane, and the coefficients of the equation of the line for each wall can be augmented in the state vector. Each ZUPT is thus an observation of a point and orientation on a line representing a wall. In our setup, we do not enforce any prior information about walls being orthogonal to each other, whereas we speculate that the RoomScan application enforces some shape constraint for the room.

\begin{figure}[!t]
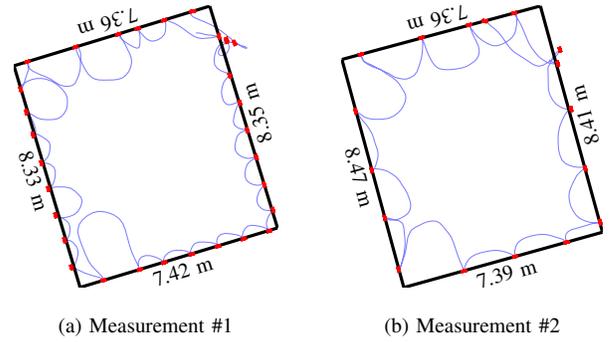


  \pgfplotsset{
    trim axis right,
    yticklabel style={rotate=90},
    grid style={very thin,gray!25}
  }

  \footnotesize\centering%
  \begin{minipage}{.49\columnwidth}

    \tikzsetnextfilename{tikz-roomscan-1}

    \setlength{\figurewidth}{1.15\columnwidth}
    \setlength{\figureheight}{0.7498\figurewidth}
    \centering
    \input{./fig/roomscan-1.tex}\\ \vspace*{1em}
    {\footnotesize {(a)~Measurement \#1}} \\
  \end{minipage}%
  \begin{minipage}{.49\columnwidth}

    \tikzsetnextfilename{tikz-roomscan-2}

    \setlength{\figurewidth}{1.15\columnwidth}
    \setlength{\figureheight}{0.7505\figurewidth}
    \centering
    \input{./fig/roomscan-2.tex}\\ \vspace*{1em}
    {\footnotesize {(b)~Measurement \#2}} \\
  \end{minipage}
  \tikzexternalenable
  \caption{Two examples of measuring the wall placements of an indoor space with an iPad Pro. The walls (the equations of the planes) are a part of the state variable. Points where the phone is stationary (against the wall) are visualized in red. The true size of the room is 7.30~m $\times$ 8.45~m.}
  \label{fig:roomscan}
  \vspace*{-1em}
\end{figure}

Markers were placed upon the walls of a room of known geometry (7.30~m $\times$ 8.45~m). The phone was moved along the walls stopping for 3~s at each marker until arriving at the starting point, such that the first two markers were visited twice.
Figure~\ref{fig:roomscan} shows the results obtained by using our model. Measurement~\#1 was done by stopping at all the available markers, while measurement~\#2 was done using only every second marker. The resulting rooms are not exactly rectangular, but remarkably close considering no orthogonality constrains were implemented. The estimated size of the room was approximately 7.4~m $\times$ 8.3~m for measurement setup~\#1 and 7.4~m $\times$  8.4~m for setup \#2. In both cases the figure shows that in the beginning of the capture the ZUPTs have not matched the wall that well, but the next observations are well matching the wall planes.

For comparison, RoomScan measurements were performed on the same markers and following a similar trajectory between them. The RoomScan application gave a rooms size of 7.3~m $\times$ 8.3~m for measurement setup~\#1 and 7.0~m $\times$ 8.6~m for setup \#2. This means that the proposed method can deliver comparable results to the black-box method implemented in RoomScan. More testing would be necessary to make an objective comparison, but the purpose of the experiment was to showcase the flexibility and potential uses of the general model presented in this paper.

\section{Discussion and Conclusions}
\label{sec:discussion}
\noindent
In this paper we have presented a general framework for inertial navigation using the limited quality data provided by standard handheld smartphones. Up till now, this has been regarded challenging, and we are not aware of any prior published work where the same would have been accomplished.

We presented a probabilistic approach building on extended Kalman filtering for continuous estimation of the position, velocity, and orientation of the mobile device. Furthermore, the IMU sensor biases and scale errors were estimated as a part of the system state. Our approach differentiates itself from prior models by directly employing the Bayesian (fully probabilistic) interpretation of non-linear state estimation (in the spirit of \cite{Sarkka:2013}), and handling the non-additive process noise inside the dynamic model. The estimation scheme avoids unnecessary approximations or error state transformations. Furthermore, we do not assume the sensor sampling rate to be fixed, but use the actual observation timestamps of the sensor events. This helps mitigate problems with missing samples and other unexpected issues with the inputs.

In order to work, the dynamic model needs to be fused with observations. We presented several types of alternative measurements that can be combined with the model. These were position fixes (see Fig.~\ref{fig:intro}), position loop-closures and barometric air pressure data (see Figs.~\ref{fig:intro} and \ref{fig:states}(b)), zero-velocity updates (all examples), and plane tangent observations (Fig.~\ref{fig:roomscan}). We also introduced constraining the speed estimate from exploding by introducing a pseudo-update for the speed. Even though many of these constraints are not general enough to fit  all applications, they still cover many potential use cases.

The presented method has many strong sides. It is general and does not requiring any steps to be detected, specific orientation to be held in, or field of vision to cover any visual features. This differentiates it from conventional PDR and odometry methods for mobile phones. The method is also not limited to estimation in a two-dimensional plane. All these aspects were covered in the experiments, where the phone was held in the pocket, in a bag, on a baby pushchair/stroller, and in an elevator. The last experiment demonstrated how the very same algorithm can be used as a measuring tool for estimating the shape and size of an indoor space.
In the PDR experiments we chose to show what the method is capable of as such. While there exists a multitude of well-tailored methods for all of the isolated test scenarios, there are no exact competing methods for mobile phones which could cover all of these scenarios. Implementing separate methods for comparison with respect to each use case was not viable, and we rather chose to put our focus on providing a convincing set of application examples.

In this paper, the data was collected using the mobile device, but the path was calculated off-line. However, the method is lightweight and capable of running in real-time on an iPhone or iPad (even older models). The computational efficiency comes from the sequential nature of the data processing, which scales linearly in the number of sensor samples.

The method still has some challenges and room for improvement. This kind of inertial navigation systems either work very well or fail miserably (\ie\ diverge)---there is no middle ground. Therefore handling of the noise scales and biases are crucial for success. Estimation of the sensor biases requires some auxiliary information to be fused with the model---be that ZUPTs, loop-closures, position fixes, or something else. Even though ZUPTs can be implemented to be performed subtly in the background (\eg\ when the user places the phone on the table), there are use cases which might be problematic. Even though, it has been argued that estimating the sensor biases as a part of the state would not be useful \cite{Nilsson+Zachariah+Skog+Handel:2013}, our experiences are quite the contrary. However, this requires the model to be both derived and implemented in a stable way avoiding unnecessary approximations in the error propagation.

The model is also sensitive to the noise scale parameters. The results in this paper benefit from the good sensors (\eg\ good dynamic range) in the Apple devices. High-end Android phones show comparable results. On Android devices the sampling rate can be set higher, which benefits the modelling (conventional strapdown INS use thousands of Hz, see \cite{Titterton+Weston:2004}).

In indoor positioning and tracking the INS presented in this paper could serve as a PDR replacement. The requirement for the zero-velocity updates could perhaps be loosened if the model would receive external position estimates based on Wi-Fi, BLE, RFID, or magnetic field anomalies.

Supplementary material for this paper available on: \\
\mbox{\url{https://aaltovision.github.io/handheld-INS/}}

\section*{Acknowledgments}
\noindent
Academy of Finland grants 277685, 295081, 308640, and 310325. We thank Manon Kok for helpful comments.

\bibliographystyle{IEEEtran}

{\small \bibliography{bibliography}}

\end{document}

%% file: fig/altitude.tex
%
%
\begin{tikzpicture}

\begin{axis}[%
width=0.951\figurewidth,
height=\figureheight,
at={(0\figurewidth,0\figureheight)},
scale only axis,
xmin=0,
xmax=120,
xtick={0,10,20,30,40,50,60,70,80,90,100,110,120},
xticklabels={{\phantom{0}},{},{},{},{},{},{},{},{},{},{},{},{\phantom{120}}},
ymin=-1,
ymax=7,
ylabel={$z$-displacement (m)},
axis background/.style={fill=white},
axis on top
]

\addplot[area legend,solid,draw=white!90!black,fill=white!90!black,forget plot]
table[row sep=crcr] {%
x	y\\
0.11	-1\\
4.6799	-1\\
4.6799	7\\
0.11	7\\
}--cycle;

\addplot[area legend,solid,draw=white!90!black,fill=white!90!black,forget plot]
table[row sep=crcr] {%
x	y\\
8.6097	-1\\
10.3196	-1\\
10.3196	7\\
8.6097	7\\
}--cycle;

\addplot[area legend,solid,draw=white!90!black,fill=white!90!black,forget plot]
table[row sep=crcr] {%
x	y\\
11.3996	-1\\
12.7295	-1\\
12.7295	7\\
11.3996	7\\
}--cycle;

\addplot[area legend,solid,draw=white!90!black,fill=white!90!black,forget plot]
table[row sep=crcr] {%
x	y\\
14.5695	-1\\
18.5394	-1\\
18.5394	7\\
14.5695	7\\
}--cycle;

\addplot[area legend,solid,draw=white!90!black,fill=white!90!black,forget plot]
table[row sep=crcr] {%
x	y\\
100.7695	-1\\
104.2096	-1\\
104.2096	7\\
100.7695	7\\
}--cycle;

\addplot[area legend,solid,draw=white!90!black,fill=white!90!black,forget plot]
table[row sep=crcr] {%
x	y\\
106.1596	-1\\
110.4496	-1\\
110.4496	7\\
106.1596	7\\
}--cycle;

\addplot[area legend,solid,draw=white!90!black,fill=white!90!black,forget plot]
table[row sep=crcr] {%
x	y\\
113.3997	-1\\
116.0098	-1\\
116.0098	7\\
113.3997	7\\
}--cycle;
\addplot [color=white!50!blue,solid,line width=1.2pt,forget plot]
  table[row sep=crcr]{%
0	-1.87975225658343e-06\\
0.1	-3.95233790056774e-06\\
0.2	9.5748990316825e-07\\
0.3	-6.90647953755608e-06\\
0.4	-7.21304864894235e-06\\
0.5	-3.45682753533453e-05\\
0.6	-6.51720871052018e-05\\
0.7	-5.88181254080637e-05\\
0.8	-3.83036671394802e-05\\
0.9	-5.09511449326168e-05\\
1	-6.55642357500169e-05\\
1.0999	-8.22475656907677e-05\\
1.1999	-0.000117099104381337\\
1.2999	-0.00015479003061075\\
1.3999	-0.000172072715830446\\
1.4999	-0.000204099466677942\\
1.5999	-0.000205151844659795\\
1.6999	-0.00021350659252649\\
1.7999	-0.000212661611210045\\
1.8999	-0.000207261158370475\\
1.9999	-0.000180823064912421\\
2.0999	-0.000193068151832662\\
2.1999	-0.000209781749895989\\
2.2999	-0.000194828726654977\\
2.3999	-0.000144321114233324\\
2.4999	-7.1851544883757e-05\\
2.5999	-9.50720120665883e-06\\
2.6999	-5.73246873714794e-06\\
2.7999	-3.43250317817387e-06\\
2.8999	7.87175162023676e-06\\
2.9999	-5.6493089472734e-07\\
3.0999	-2.02990026449737e-05\\
3.1999	-3.78395685395913e-05\\
3.2999	-7.8021965137486e-05\\
3.3999	-0.000138170375082389\\
3.4999	-0.00021578201955849\\
3.5999	-0.000328689082951419\\
3.6999	-0.000455014147425631\\
3.7999	-0.000569488851796345\\
3.8999	-0.000659591596221743\\
3.9999	-0.000704345080559703\\
4.0999	-0.000720897287555995\\
4.1999	-0.000736769584172472\\
4.2999	-0.000746287717177177\\
4.3999	-0.00074095419822411\\
4.4999	-0.00074117024347005\\
4.5999	-0.000812971982702793\\
4.6999	-0.00113683786094241\\
4.7999	-0.00175321891360218\\
4.8998	-0.00262445543757096\\
4.9998	-0.00371652329061227\\
5.0998	-0.00508834467184851\\
5.1998	-0.00666802414318113\\
5.2998	-0.00847014012987658\\
5.3998	-0.010538395100638\\
5.4997	-0.0128470306189879\\
5.5997	-0.0153627473894661\\
5.6998	-0.0180830484928482\\
5.7998	-0.0210525099254649\\
5.8998	-0.0241705526802297\\
5.9998	-0.0275326112694747\\
6.0997	-0.031099250181402\\
6.1997	-0.0347656089792937\\
6.2997	-0.0379581152918918\\
6.3997	-0.0393196803855779\\
6.4997	-0.0285301247471528\\
6.5997	0.0182694140371341\\
6.6997	0.109744558862144\\
6.7997	0.246930744682299\\
6.8997	0.412732031604365\\
6.9997	0.574536762752511\\
7.0997	0.723106040634789\\
7.1997	0.839395116182898\\
7.2997	0.927559800186816\\
7.3997	0.974373063759743\\
7.4997	0.985223755421538\\
7.5997	0.998318290443318\\
7.6997	1.01017938626672\\
7.7997	1.01978454840261\\
7.8997	1.02416896580155\\
7.9997	1.024783708764\\
8.0997	1.02237433997482\\
8.1997	1.01910918701042\\
8.2997	1.01743509782342\\
8.3997	1.01742481043725\\
8.4997	1.0163350670534\\
8.5997	1.01588066761827\\
8.6997	1.01585768640149\\
8.7997	1.01582281056801\\
8.8997	1.01580404204294\\
8.9997	1.0157771113334\\
9.0997	1.01573325683623\\
9.1997	1.01566475308417\\
9.2997	1.01564283220598\\
9.3997	1.01560166327051\\
9.4997	1.01558456866448\\
9.5997	1.01558751963652\\
9.6997	1.01557440148523\\
9.7997	1.01556173099236\\
9.8997	1.01554590168986\\
9.9996	1.01553137266157\\
10.0996	1.01552355457574\\
10.1996	1.01552111589466\\
10.2996	1.01551849518833\\
10.3996	1.01552535766073\\
10.4996	1.01552233671574\\
10.5996	1.01547036511497\\
10.6996	1.01542213643605\\
10.7996	1.01531693530728\\
10.8996	1.01519274912691\\
10.9996	1.0150937380637\\
11.0996	1.01500282447666\\
11.1996	1.01490238375018\\
11.2996	1.01483866393004\\
11.3996	1.01481992823591\\
11.4995	1.01479344660282\\
11.5995	1.01479726094326\\
11.6996	1.01478887011548\\
11.7996	1.01475069143543\\
11.8995	1.01469850021115\\
11.9996	1.01468159882894\\
12.0995	1.01465189070953\\
12.1995	1.01463538444617\\
12.2996	1.01462982609084\\
12.3995	1.01460927609317\\
12.4995	1.01460169151861\\
12.5995	1.01462247910838\\
12.6995	1.0146375799397\\
12.7995	1.01474390595975\\
12.8995	1.01509350791338\\
12.9995	1.01570312564199\\
13.0995	1.01656595797055\\
13.1995	1.01764523298226\\
13.2995	1.018948039789\\
13.3995	1.02053189933885\\
13.4995	1.02258228086969\\
13.5995	1.02401376005534\\
13.6995	1.01609344432303\\
13.7995	1.00315316957334\\
13.8995	0.995255914720928\\
13.9995	0.989061703243052\\
14.0995	0.986236037954068\\
14.1995	0.985005042216415\\
14.2995	0.986176426160534\\
14.3995	0.987639554369184\\
14.4995	0.988458042310904\\
14.5995	0.98855878963824\\
14.6995	0.988555965105578\\
14.7995	0.988536575148237\\
14.8995	0.988532965197864\\
14.9994	0.98852046917976\\
15.0994	0.988520202060814\\
15.1994	0.988513691921095\\
15.2994	0.988508922854116\\
15.3994	0.988506049305301\\
15.4994	0.98849206470653\\
15.5994	0.988502117451073\\
15.6994	0.988489746631592\\
15.7994	0.988486753178019\\
15.8994	0.988489418071113\\
15.9994	0.988488631371618\\
16.0994	0.988472130380593\\
16.1994	0.988471835696868\\
16.2994	0.988475679770146\\
16.3994	0.988475688950029\\
16.4994	0.988459484000047\\
16.5994	0.988450030323153\\
16.6994	0.988448374734263\\
16.7994	0.988443733641026\\
16.8994	0.988445335243886\\
16.9994	0.988449233869707\\
17.0994	0.98847951297243\\
17.1994	0.988505110512333\\
17.2994	0.988515161561656\\
17.3994	0.988507986054376\\
17.4994	0.988482920939916\\
17.5994	0.988459059707262\\
17.6994	0.988418925125005\\
17.7994	0.988422751668284\\
17.8994	0.988449446284792\\
17.9994	0.988480611729569\\
18.0994	0.988491666326431\\
18.1994	0.988495594777787\\
18.2994	0.988504053605904\\
18.3994	0.988521022472122\\
18.4994	0.98853699858554\\
18.5994	0.988604974751197\\
18.6994	0.98859268126929\\
18.7994	0.988067172672462\\
18.8994	0.986683015920979\\
18.9994	0.989424479085829\\
19.0994	0.996233350164033\\
19.1994	1.00510249098586\\
19.2994	1.0174351981846\\
19.3994	1.03389825077305\\
19.4994	1.05072208919096\\
19.5994	1.06148788351807\\
19.6994	1.06582087783733\\
19.7994	1.06548217466511\\
19.8994	1.0645147454993\\
19.9993	1.06279094341134\\
20.0993	1.06508669083458\\
20.1993	1.06650820775136\\
20.2993	1.06792536758075\\
20.3993	1.06284230203798\\
20.4993	1.05312642359881\\
20.5993	1.05246915027852\\
20.6993	1.06413693748897\\
20.7993	1.07220860264207\\
20.8993	1.07270651491997\\
20.9993	1.0650964251478\\
21.0993	1.0543643994484\\
21.1993	1.05323670485637\\
21.2993	1.06806610961822\\
21.3993	1.08500278370902\\
21.4993	1.09443743441376\\
21.5993	1.08765530016493\\
21.6993	1.07713309449889\\
21.7993	1.08367052203337\\
21.8993	1.10609250775922\\
21.9993	1.11556108035798\\
22.0993	1.10432015984173\\
22.1993	1.08310676923316\\
22.2993	1.08005826902109\\
22.3993	1.10084235801592\\
22.4993	1.12237456821006\\
22.5993	1.1215597039139\\
22.6993	1.1021009748202\\
22.7993	1.0908910048719\\
22.8993	1.1046164305209\\
22.9993	1.13006836130489\\
23.0993	1.1273337783992\\
23.1993	1.10774276977415\\
23.2993	1.10354831293744\\
23.3993	1.12087903726385\\
23.4993	1.13547297986603\\
23.5993	1.12958934488348\\
23.6993	1.11082602616383\\
23.7993	1.1049956328475\\
23.8993	1.12235533204173\\
23.9993	1.13960949875559\\
24.0993	1.13257186518345\\
24.1993	1.1203226715375\\
24.2993	1.1300108745208\\
24.3993	1.15479059514184\\
24.4993	1.16799563721001\\
24.5993	1.1601653629698\\
24.6993	1.14355720508349\\
24.7993	1.14351180009837\\
24.8993	1.16212763001138\\
24.9992	1.17113238133369\\
25.0992	1.15707620361935\\
25.1992	1.14500647688248\\
25.2992	1.15378172901527\\
25.3992	1.16660326553334\\
25.4992	1.16986358386197\\
25.5992	1.15972676619182\\
25.6992	1.14392949610498\\
25.7992	1.14867495241087\\
25.8992	1.17105053394885\\
25.9992	1.17588062257325\\
26.0992	1.15759471028811\\
26.1992	1.15095227637972\\
26.2992	1.16763220333553\\
26.3992	1.18802426089168\\
26.4992	1.19085860826031\\
26.5992	1.17505063752381\\
26.6992	1.16495203729463\\
26.7992	1.1819156569447\\
26.8992	1.20444557961298\\
26.9992	1.20161562138112\\
27.0992	1.18043154383729\\
27.1992	1.17036755028715\\
27.2992	1.18712942804701\\
27.3992	1.20560059937511\\
27.4992	1.20567764243563\\
27.5992	1.18755006312254\\
27.6992	1.17907459444784\\
27.7992	1.19981999981316\\
27.8992	1.23060893036645\\
27.9992	1.23604712210188\\
28.0992	1.21320236340291\\
28.1992	1.19589917185957\\
28.2992	1.20822935196673\\
28.3992	1.23316346090777\\
28.4992	1.24238537771997\\
28.5992	1.22631609806485\\
28.6992	1.20792776670565\\
28.7992	1.2127080674048\\
28.8992	1.24405915193153\\
28.9992	1.26495454487593\\
29.0992	1.26602066713471\\
29.1992	1.26306983806566\\
29.2992	1.28045182975407\\
29.3992	1.31933916678032\\
29.4992	1.35710073050163\\
29.5992	1.3847083664759\\
29.6992	1.4086651306106\\
29.7992	1.44497023735738\\
29.8992	1.50338931243886\\
29.9991	1.54454650067393\\
30.0991	1.5675664652082\\
30.1991	1.59100912944806\\
30.2991	1.63686274860906\\
30.3991	1.69142263589293\\
30.4991	1.72746364451086\\
30.5991	1.74573787969188\\
30.6991	1.77588656458745\\
30.7991	1.82841034673777\\
30.8991	1.87988520483344\\
30.9991	1.91318789697416\\
31.0991	1.94016976762215\\
31.1991	1.98496048533192\\
31.2991	2.03654943735012\\
31.3991	2.07749137213539\\
31.4991	2.09872345971944\\
31.5991	2.12670440754931\\
31.6991	2.17401916506129\\
31.7991	2.22421857930937\\
31.8991	2.25692444837568\\
31.9991	2.27549327556317\\
32.0991	2.30575909220172\\
32.1991	2.35868612534274\\
32.2991	2.40553510171573\\
32.3991	2.43585402619744\\
32.4991	2.45251172735864\\
32.5991	2.48411232610227\\
32.6991	2.53991355725278\\
32.7991	2.59600066690024\\
32.8991	2.63033901934116\\
32.9991	2.63790098337009\\
33.0991	2.6402605109386\\
33.1991	2.66261303568704\\
33.2991	2.7096368867681\\
33.3991	2.75372855616966\\
33.4991	2.79009150751617\\
33.5991	2.81000087749872\\
33.6991	2.81552478654929\\
33.7991	2.84436540806099\\
33.8991	2.91374573499573\\
33.9991	2.9726503059199\\
34.0991	3.00803557126545\\
34.1991	3.02483300406141\\
34.2991	3.05114238959872\\
34.3991	3.09993458641396\\
34.4991	3.14509279806364\\
34.5991	3.17317075929253\\
34.6991	3.19387555405599\\
34.7991	3.22333468916196\\
34.8991	3.28130020686098\\
34.9991	3.33672236627109\\
35.0991	3.3687814153004\\
35.1991	3.38416869629049\\
35.2991	3.41705095551872\\
35.3991	3.47948327773926\\
35.4991	3.53316983838142\\
35.5991	3.56609907644365\\
35.6991	3.5984483553851\\
35.7991	3.65500699299043\\
35.8991	3.72072876669568\\
35.9991	3.75654931517578\\
36.0991	3.77046118754559\\
36.1991	3.79141263877228\\
36.2991	3.84303614431632\\
36.3991	3.89478601325352\\
36.4991	3.9292044970054\\
36.5991	3.94814607783793\\
36.6991	3.97913982144182\\
36.7991	4.03813855518044\\
36.8991	4.10121907396275\\
36.9991	4.13574657720063\\
37.0991	4.15370900659873\\
37.1991	4.1866855122732\\
37.2991	4.23704169872066\\
37.3991	4.2780174159012\\
37.4991	4.30364442954554\\
37.5991	4.32635879044086\\
37.6991	4.37794799746636\\
37.7991	4.44740767860282\\
37.8991	4.49399228635786\\
37.9991	4.51608160408674\\
38.0991	4.53792937215744\\
38.1991	4.58445218215251\\
38.2991	4.63503188883567\\
38.3991	4.66691630350443\\
38.4991	4.68656200275492\\
38.5991	4.71681418933862\\
38.6991	4.77454905470059\\
38.7991	4.84691167142906\\
38.8991	4.89210881663631\\
38.9991	4.91303455627479\\
39.0991	4.93156155045823\\
39.1991	4.97981550408897\\
39.2991	5.03307240830369\\
39.3991	5.0728217136647\\
39.4991	5.09196464999923\\
39.5991	5.11895259200473\\
39.6991	5.17412991075279\\
39.7991	5.23589161271382\\
39.8991	5.27266384064064\\
39.9991	5.29570298731872\\
40.0991	5.33059466154758\\
40.1991	5.38823015282702\\
40.2991	5.4384245135714\\
40.3991	5.47364458338295\\
40.499	5.49878011677992\\
40.5991	5.54291176238629\\
40.6991	5.61439805316901\\
40.7991	5.67472220735379\\
40.8991	5.71452671846159\\
40.9991	5.73789712341395\\
41.0991	5.77589485074953\\
41.1991	5.83788177343553\\
41.2991	5.894596284582\\
41.399	5.93534027702684\\
41.4991	5.95488084945047\\
41.5991	5.96256446576681\\
41.6991	5.98700437896173\\
41.7991	6.04770439443058\\
41.8991	6.10515500940704\\
41.999	6.1388789253592\\
42.0991	6.14380079318515\\
42.1991	6.13318332987053\\
42.2991	6.13230682649418\\
42.3991	6.15855786695548\\
42.4991	6.18622192954967\\
42.5991	6.20527249821477\\
42.6991	6.20870782456456\\
42.7991	6.20528835349744\\
42.8991	6.21015541826612\\
42.9991	6.23170622621011\\
43.0991	6.25590538057425\\
43.1991	6.25914634244849\\
43.2991	6.25155764670323\\
43.3991	6.24768281098106\\
43.4991	6.26047507541726\\
43.5991	6.28112686710642\\
43.6991	6.29983313438619\\
43.7991	6.30958825729462\\
43.8991	6.31202784246618\\
43.9991	6.32059145013364\\
44.0991	6.33937091264305\\
44.1991	6.35487935560896\\
44.2991	6.35438735850128\\
44.3991	6.34234605555139\\
44.4991	6.32613311467161\\
44.5991	6.31207524068159\\
44.6991	6.29500905450829\\
44.7991	6.280029941485\\
44.8991	6.26168738097011\\
44.999	6.2371395444342\\
45.099	6.20881125991807\\
45.199	6.19331960258703\\
45.299	6.18158843861412\\
45.399	6.1628159579644\\
45.499	6.13292616728073\\
45.599	6.10252774814818\\
45.699	6.08330511803042\\
45.799	6.0734610309098\\
45.899	6.06958275004735\\
45.999	6.05588511767612\\
46.099	6.02438470714001\\
46.199	5.99829373467508\\
46.299	5.98109235913359\\
46.399	5.98415713305247\\
46.499	5.97727723709491\\
46.599	5.94740078638223\\
46.699	5.91281251977713\\
46.799	5.89059286183862\\
46.899	5.86377992423236\\
46.999	5.83311585162598\\
47.099	5.79656026702734\\
47.199	5.74958448308241\\
47.299	5.73060397641962\\
47.399	5.7194881346176\\
47.499	5.72283535322697\\
47.599	5.72831027665843\\
47.699	5.70649727067935\\
47.799	5.69548475402381\\
47.899	5.68738940006194\\
47.999	5.6800393858493\\
48.099	5.66158355883033\\
48.199	5.62941924277735\\
48.299	5.60849324535944\\
48.399	5.60926864070637\\
48.4991	5.63088417052095\\
48.599	5.64390577819877\\
48.6991	5.63734198802203\\
48.7991	5.61403092581097\\
48.899	5.60462174112081\\
48.9991	5.62356179858617\\
49.0991	5.6439733602104\\
49.1991	5.64067082900929\\
49.2991	5.62504990737452\\
49.3991	5.6252719281696\\
49.499	5.6540461378781\\
49.5991	5.67998205698406\\
49.6991	5.67567854374324\\
49.7991	5.64252941821029\\
49.8991	5.61332453639043\\
49.999	5.61797701710537\\
50.099	5.63407820172447\\
50.199	5.63705657420417\\
50.299	5.60895031768225\\
50.399	5.58390793231065\\
50.499	5.59759421989245\\
50.599	5.62701348026682\\
50.699	5.64090995706021\\
50.799	5.62570882061578\\
50.899	5.60405610002698\\
50.999	5.6123968580355\\
51.099	5.64640249308634\\
51.199	5.67311410591465\\
51.299	5.67130225075249\\
51.399	5.65260004915592\\
51.499	5.6470087195636\\
51.599	5.67205794397709\\
51.699	5.69594271285828\\
51.799	5.69564674677136\\
51.899	5.66902306781478\\
51.999	5.64937103313872\\
52.099	5.66830146540463\\
52.199	5.70161415635398\\
52.299	5.7150356657989\\
52.399	5.704425031929\\
52.499	5.69232266269771\\
52.599	5.70589120354809\\
52.699	5.72676774301318\\
52.799	5.7312372806327\\
52.899	5.70898770424335\\
52.999	5.68751573482855\\
53.099	5.69119039411791\\
53.199	5.71535497843585\\
53.2991	5.73430955658394\\
53.3991	5.72764167483611\\
53.4991	5.71732525504175\\
53.5991	5.72334007498784\\
53.6991	5.74860139824117\\
53.7991	5.76560731074258\\
53.8991	5.76010456522758\\
53.999	5.74426004451496\\
54.0991	5.74294257852199\\
54.1991	5.77245018251229\\
54.299	5.80877487493622\\
54.3991	5.81736276701522\\
54.4991	5.80269456730148\\
54.599	5.78512229828477\\
54.6991	5.79890413036834\\
54.7991	5.82166481597054\\
54.8991	5.83722083365636\\
54.999	5.83778772762765\\
55.099	5.84622590235572\\
55.199	5.87871956112129\\
55.299	5.93055639758517\\
55.399	5.973759134858\\
55.499	5.99820497900944\\
55.599	6.02677673814792\\
55.699	6.05570548435146\\
55.799	6.09312959980215\\
55.899	6.13340778427858\\
55.999	6.16386641008113\\
56.099	6.18045896100801\\
56.199	6.19501356886504\\
56.299	6.20991877830906\\
56.399	6.22948817847612\\
56.499	6.2377737773344\\
56.599	6.22808757722736\\
56.699	6.22884050463363\\
56.799	6.25664138349061\\
56.899	6.29560380034235\\
56.999	6.31217429844381\\
57.099	6.3057097537028\\
57.199	6.29274324098653\\
57.299	6.29968200413777\\
57.399	6.3206706163166\\
57.4991	6.33440768947044\\
57.5991	6.33050463440568\\
57.6991	6.33290724478882\\
57.7991	6.35546911229082\\
57.8991	6.41118732620308\\
57.9991	6.47141623569708\\
58.0991	6.50161895048845\\
58.1991	6.50213625109927\\
58.2991	6.50285169403688\\
58.399	6.53045712616126\\
58.4991	6.57739958806292\\
58.5991	6.60556731342955\\
58.6991	6.59882425709332\\
58.7991	6.5851751087076\\
58.8991	6.59528923242464\\
58.9991	6.60864760539422\\
59.0991	6.61592434420667\\
59.1991	6.60532120610696\\
59.2991	6.58613755917429\\
59.3991	6.59382621905267\\
59.4991	6.60882456060149\\
59.5991	6.61075564957963\\
59.6991	6.59695065253572\\
59.7991	6.57749512442724\\
59.8991	6.57738146598899\\
59.999	6.58416228527372\\
60.099	6.59366201970303\\
60.199	6.58152086129073\\
60.299	6.54788182529704\\
60.399	6.53927918465832\\
60.499	6.55304010574823\\
60.599	6.56170251818758\\
60.699	6.55363200219301\\
60.799	6.52834972386207\\
60.899	6.51352487137652\\
60.999	6.52127571776708\\
61.099	6.55167703552514\\
61.199	6.56031613447699\\
61.299	6.52869728716875\\
61.399	6.50711082603525\\
61.499	6.52371151039488\\
61.599	6.53456273730568\\
61.699	6.53200288465383\\
61.799	6.50841710647714\\
61.899	6.48285990789117\\
61.999	6.4802831761874\\
62.099	6.50669876454306\\
62.199	6.52811687584546\\
62.299	6.50156686788208\\
62.399	6.47346790709107\\
62.499	6.48082530688247\\
62.599	6.49335642590519\\
62.699	6.49604129537234\\
62.799	6.47538009705581\\
62.899	6.44404467265589\\
62.999	6.43388636608436\\
63.099	6.44894210566466\\
63.199	6.47821249589278\\
63.299	6.47189661117326\\
63.399	6.4340806652726\\
63.499	6.43009504907627\\
63.599	6.44589075642159\\
63.699	6.45160413864606\\
63.799	6.43659514546618\\
63.8991	6.40483738252572\\
63.999	6.38333256508367\\
64.099	6.38845033618503\\
64.1991	6.41693884059673\\
64.2991	6.42257108865889\\
64.3991	6.38702469561529\\
64.4991	6.36674300408181\\
64.599	6.38054207366718\\
64.6991	6.38757650715508\\
64.7991	6.38079008247882\\
64.8991	6.35195535798283\\
64.999	6.32077346888356\\
65.099	6.31325157925614\\
65.199	6.33506264453298\\
65.299	6.35430373698031\\
65.399	6.32835118300011\\
65.499	6.29672765928344\\
65.599	6.30384176975217\\
65.699	6.31278930655489\\
65.799	6.31122253025044\\
65.899	6.28807305223758\\
65.999	6.25393296440783\\
66.099	6.24117748969337\\
66.199	6.25452941250189\\
66.299	6.279609921061\\
66.399	6.26703416330723\\
66.499	6.22826646327333\\
66.599	6.22520904427097\\
66.699	6.23610887320306\\
66.799	6.23878475267362\\
66.899	6.2312591308252\\
66.999	6.20721422785565\\
67.099	6.19037671887932\\
67.1991	6.20752007352789\\
67.299	6.24652856334967\\
67.3991	6.26969677322022\\
67.4991	6.25978990634225\\
67.5991	6.23567998919315\\
67.6991	6.22681677856657\\
67.7991	6.25246406428974\\
67.8991	6.30235841561492\\
67.9991	6.34595771652266\\
68.0991	6.36248616498254\\
68.1991	6.37094082522274\\
68.2991	6.38268008794714\\
68.3991	6.3850224532071\\
68.4991	6.37124208909096\\
68.5991	6.34779569893624\\
68.6991	6.33297417657656\\
68.7991	6.33005091367788\\
68.8991	6.33076050568373\\
68.9991	6.32387198246379\\
69.0991	6.30065397172439\\
69.1991	6.27424004212748\\
69.2991	6.26547490745406\\
69.3991	6.26688291349126\\
69.4991	6.25588672723067\\
69.5991	6.22734933834159\\
69.6991	6.19859092016921\\
69.7991	6.18734580729592\\
69.8991	6.18924420444853\\
69.999	6.18536320685114\\
70.099	6.16890442772727\\
70.199	6.14785377103694\\
70.2991	6.13993945326816\\
70.3991	6.14160490338042\\
70.4991	6.13436979203892\\
70.599	6.11251310972244\\
70.699	6.08822464499977\\
70.7991	6.07454309728165\\
70.8991	6.07780301223333\\
70.9991	6.08612591430268\\
71.0991	6.08208586332644\\
71.1991	6.0612299316412\\
71.2991	6.04315866385975\\
71.3991	6.04272793067952\\
71.4991	6.0497697479957\\
71.5991	6.04708346837336\\
71.6991	6.02919937066548\\
71.7991	6.01323685167784\\
71.8991	6.01218052721842\\
71.9991	6.02517941002898\\
72.0991	6.03541972913516\\
72.1991	6.03055461655173\\
72.2991	6.01416006870189\\
72.3991	6.00489149251115\\
72.4991	6.02044491950657\\
72.5991	6.03468330950279\\
72.6991	6.03304078696805\\
72.7991	6.01390107858152\\
72.8991	6.00212900372068\\
72.9991	6.00944920651704\\
73.0991	6.02223658614661\\
73.1991	6.03321475237159\\
73.2991	6.03216333390021\\
73.3991	6.02551342826131\\
73.4991	6.01757002047852\\
73.5991	6.02005056357744\\
73.6991	6.01736204596707\\
73.7991	6.00554966558011\\
73.8991	5.96963408660062\\
73.9991	5.90385660578694\\
74.0991	5.84108466583577\\
74.1991	5.80366112091963\\
74.2991	5.78717506587476\\
74.3991	5.76030064780778\\
74.4991	5.70109128221431\\
74.5991	5.63791780885373\\
74.6991	5.59224458135761\\
74.7991	5.5606596680233\\
74.8991	5.50879864863271\\
74.9991	5.44317986537535\\
75.099	5.39672348859399\\
75.1991	5.37333528086231\\
75.2991	5.34636855542766\\
75.3991	5.29707693145028\\
75.4991	5.23886949154155\\
75.5991	5.19819460781429\\
75.6991	5.16623232780569\\
75.7991	5.1179513778446\\
75.8991	5.05824227514389\\
75.9991	5.01384279573999\\
76.0991	4.99062345427741\\
76.1991	4.96493577261551\\
76.2991	4.91488445274296\\
76.3991	4.86350022585841\\
76.4991	4.82663464112831\\
76.5991	4.796390851505\\
76.6991	4.74731991927018\\
76.7991	4.69347266712468\\
76.8991	4.64690652002797\\
76.9991	4.61495500069803\\
77.0991	4.58014993133977\\
77.1991	4.52853007345135\\
77.2991	4.48111100374678\\
77.3991	4.44770298767615\\
77.4991	4.41839462526251\\
77.5991	4.36793151435139\\
77.6991	4.31798728466593\\
77.7991	4.29064226092557\\
77.8991	4.26372792140273\\
77.9991	4.22062092204316\\
78.0991	4.1692196303714\\
78.1991	4.1255131916208\\
78.2991	4.09338384789571\\
78.3991	4.05800293277525\\
78.4991	3.99735443725046\\
78.5991	3.94701255260017\\
78.6991	3.91285212205618\\
78.7991	3.87568345675142\\
78.8991	3.8295354071878\\
78.9991	3.7846310528354\\
79.0991	3.7522357443725\\
79.1992	3.72182471251812\\
79.2992	3.67352276123046\\
79.3992	3.61119613141033\\
79.4992	3.57256763871261\\
79.5991	3.55034693489514\\
79.6992	3.52342217701273\\
79.7992	3.47179505591514\\
79.8992	3.42993684536733\\
79.9991	3.38825061850389\\
80.0991	3.34808995525574\\
80.1991	3.28872081360728\\
80.2991	3.24279923521175\\
80.3991	3.20523861146103\\
80.4991	3.17879418676442\\
80.5992	3.13360483404801\\
80.6992	3.08627649727287\\
80.7991	3.04797045006443\\
80.8991	3.01240746923984\\
80.9992	2.96205445428549\\
81.0992	2.91927628891871\\
81.1992	2.88912645849085\\
81.2992	2.87341684605418\\
81.3991	2.85539569576377\\
81.4992	2.83422657350177\\
81.5992	2.83304729007637\\
81.6992	2.85229459169481\\
81.7992	2.86590474684357\\
81.8992	2.86601368526278\\
81.9992	2.86067958564585\\
82.0992	2.85563522939899\\
82.1992	2.85859955111449\\
82.2992	2.85660947899623\\
82.3992	2.84604773010826\\
82.4992	2.80707035876503\\
82.5992	2.74487854584152\\
82.6992	2.69234383440228\\
82.7992	2.66696216509651\\
82.8992	2.63883350869331\\
82.9992	2.57998262607165\\
83.0992	2.50868830548029\\
83.1992	2.45446009208683\\
83.2992	2.4224231996974\\
83.3992	2.37846941543114\\
83.4992	2.30371509347809\\
83.5992	2.23510146619762\\
83.6992	2.20694284299775\\
83.7992	2.19453597077968\\
83.8992	2.15365486998901\\
83.9992	2.09312139735044\\
84.0992	2.04644444425058\\
84.1992	2.04280368650679\\
84.2992	2.03249969170637\\
84.3992	1.98607956180968\\
84.4992	1.92867722975531\\
84.5993	1.88072482213687\\
84.6992	1.84432658148978\\
84.7993	1.78379228295364\\
84.8993	1.72648043691729\\
84.9991	1.69213071642096\\
85.0991	1.67240271895535\\
85.1992	1.63183233012803\\
85.2992	1.58071338254755\\
85.3992	1.52973847758994\\
85.4992	1.49826323015588\\
85.5992	1.45696652765988\\
85.6992	1.40803115240256\\
85.7992	1.36536113777363\\
85.8992	1.3327915893903\\
85.9992	1.29955750810168\\
86.0992	1.24831910898889\\
86.1992	1.21291518947382\\
86.2992	1.2160041846716\\
86.3992	1.22397321564499\\
86.4992	1.20820892363946\\
86.5992	1.18752603052223\\
86.6992	1.18947891060615\\
86.7992	1.21362444371667\\
86.8992	1.23070753777337\\
86.9992	1.23016534271755\\
87.0992	1.21514818994786\\
87.1992	1.20719100197666\\
87.2992	1.22947281231026\\
87.3992	1.25588377668982\\
87.4992	1.2523480753131\\
87.5992	1.22785473884503\\
87.6992	1.21490499640408\\
87.7992	1.22874661795673\\
87.8992	1.24631812192033\\
87.9992	1.24597063458863\\
88.0992	1.23183332471009\\
88.1992	1.21938174780441\\
88.2992	1.22309253303561\\
88.3993	1.23628615271362\\
88.4993	1.22905015389054\\
88.5992	1.20322071496083\\
88.6993	1.18351497895381\\
88.7993	1.18776612603865\\
88.8993	1.20533586317549\\
88.9993	1.20896856618053\\
89.0993	1.19334556310163\\
89.1993	1.17015302103818\\
89.2993	1.16133547241127\\
89.3993	1.17342131101055\\
89.4993	1.17420237676001\\
89.5993	1.16022057204152\\
89.6993	1.14147853306593\\
89.7993	1.13641375601575\\
89.8993	1.14817762171498\\
89.9993	1.16175489433148\\
90.0993	1.15966082267573\\
90.1993	1.14790439342741\\
90.2993	1.13610079169894\\
90.3993	1.13946170605563\\
90.4993	1.15129218840218\\
90.5993	1.14701579787819\\
90.6993	1.12907931916985\\
90.7993	1.11415405995576\\
90.8993	1.11317683169437\\
90.9993	1.12752556933436\\
91.0993	1.1403946963163\\
91.1993	1.1416959575915\\
91.2993	1.13111412897473\\
91.3993	1.12672500080029\\
91.4993	1.14328690630128\\
91.5993	1.15847387599806\\
91.6993	1.15116932463245\\
91.7993	1.1303325468827\\
91.8993	1.1155181303447\\
91.9993	1.11774097371043\\
92.0993	1.12604094705761\\
92.1993	1.12998485701329\\
92.2993	1.11894477391606\\
92.3993	1.10308936684464\\
92.4993	1.10215375179839\\
92.5993	1.12429038742098\\
92.6993	1.13498105968864\\
92.7993	1.12452600954576\\
92.8993	1.10566065323806\\
92.9994	1.0994238137279\\
93.0993	1.11473553557376\\
93.1994	1.13053458916793\\
93.2994	1.12938969067675\\
93.3994	1.1175868102603\\
93.4994	1.11019125135986\\
93.5994	1.1175808393095\\
93.6994	1.12657718668332\\
93.7994	1.12301927167329\\
93.8994	1.10584082103538\\
93.9994	1.09543206448299\\
94.0994	1.10656423995846\\
94.1994	1.12835218499736\\
94.2994	1.14170083000552\\
94.3994	1.13257421903456\\
94.4994	1.11423575238418\\
94.5994	1.10761108806818\\
94.6994	1.12144688789936\\
94.7994	1.12621810356654\\
94.8994	1.11633371445261\\
94.9993	1.09474427488225\\
95.0993	1.07185385611318\\
95.1993	1.05305655596526\\
95.2994	1.02685313064688\\
95.3994	0.992403535937593\\
95.4994	0.946990592043059\\
95.5994	0.905193115092062\\
95.6994	0.888101245846062\\
95.7994	0.885342470697922\\
95.8994	0.87821515405646\\
95.9994	0.861466484817101\\
96.0994	0.850008667397665\\
96.1994	0.85819582041099\\
96.2994	0.888209879840735\\
96.3994	0.913423187232523\\
96.4994	0.92307805911983\\
96.5994	0.937041187557684\\
96.6994	0.970006361041794\\
96.7994	1.00539692906697\\
96.8994	1.0172845839621\\
96.9994	1.01198349282275\\
97.0994	1.0158162628329\\
97.1994	1.03051778130726\\
97.2994	1.05453459664914\\
97.3994	1.05704607325461\\
97.4994	1.04457232048421\\
97.5994	1.03728035302095\\
97.6994	1.04936090822296\\
97.7994	1.06264740807494\\
97.8994	1.0622013397189\\
97.9994	1.05008246931524\\
98.0994	1.04962012844897\\
98.1995	1.05989586293789\\
98.2995	1.07476644048911\\
98.3995	1.07636384378572\\
98.4995	1.06622203841892\\
98.5995	1.04907828217361\\
98.6995	1.04543288463163\\
98.7995	1.05936910017988\\
98.8995	1.07854090544593\\
98.9995	1.09208083092521\\
99.0995	1.10200480591715\\
99.1995	1.11880801234574\\
99.2995	1.14057340563755\\
99.3995	1.15750191288773\\
99.4995	1.16220085891863\\
99.5995	1.16028942076529\\
99.6995	1.15737932798067\\
99.7995	1.15618757388502\\
99.8995	1.15544738467589\\
99.9994	1.15208932261284\\
100.0994	1.1511700811166\\
100.1995	1.15006951459578\\
100.2995	1.1501397907843\\
100.3995	1.14972373636243\\
100.4995	1.14973259076023\\
100.5995	1.14927816366122\\
100.6995	1.1493842225252\\
100.7995	1.14950807808685\\
100.8995	1.14948554283531\\
100.9995	1.14947818596541\\
101.0995	1.14948544129942\\
101.1995	1.14950208597568\\
101.2995	1.14950582252439\\
101.3995	1.14948767174245\\
101.4995	1.14946123728883\\
101.5995	1.14945898093058\\
101.6995	1.14945529202646\\
101.7995	1.14949269948976\\
101.8995	1.14949974708042\\
101.9995	1.14947357972985\\
102.0995	1.14944859711973\\
102.1995	1.14943053828264\\
102.2995	1.14939385964618\\
102.3995	1.14939099825663\\
102.4995	1.14936345514095\\
102.5995	1.14934156434535\\
102.6995	1.14936631125493\\
102.7995	1.14942117928113\\
102.8995	1.14944780869392\\
102.9995	1.149450004182\\
103.0995	1.14943634610984\\
103.1996	1.14942971512472\\
103.2996	1.14943132276337\\
103.3996	1.14946672714376\\
103.4996	1.14948455951632\\
103.5996	1.14947335914761\\
103.6996	1.14944814740188\\
103.7996	1.14943790605388\\
103.8996	1.14945475628895\\
103.9996	1.14945209950052\\
104.0996	1.14952070903089\\
104.1996	1.1499289029925\\
104.2996	1.15062886818305\\
104.3996	1.1508287077097\\
104.4996	1.15257845399078\\
104.5996	1.15612191864239\\
104.6996	1.14742937055719\\
104.7996	1.1369705141513\\
104.8996	1.13326389083592\\
104.9996	1.13478754282704\\
105.0996	1.13677728807296\\
105.1996	1.1398763529937\\
105.2996	1.14006609310183\\
105.3996	1.14063051594144\\
105.4996	1.14093332346937\\
105.5996	1.14099367574982\\
105.6996	1.14089717761639\\
105.7996	1.14077941627267\\
105.8996	1.14066008350389\\
105.9996	1.14059159233387\\
106.0996	1.14053164136457\\
106.1996	1.14051923441015\\
106.2996	1.14052757294746\\
106.3996	1.14051570252148\\
106.4996	1.14051861295625\\
106.5996	1.14054923596298\\
106.6996	1.14055654593638\\
106.7996	1.14055385275403\\
106.8996	1.14056592371083\\
106.9996	1.14056484976848\\
107.0996	1.14056144618709\\
107.1996	1.14056580203786\\
107.2996	1.14058112826812\\
107.3996	1.14058015218956\\
107.4996	1.14055703672438\\
107.5996	1.14054715718996\\
107.6996	1.14055824150546\\
107.7996	1.14055592958752\\
107.8996	1.14056355576479\\
107.9996	1.14058589510019\\
108.0996	1.14057162050838\\
108.1997	1.14057427343549\\
108.2996	1.14058614764871\\
108.3997	1.14058292088614\\
108.4997	1.14057944735934\\
108.5997	1.14058229203155\\
108.6997	1.14056787781624\\
108.7997	1.14056661657999\\
108.8997	1.14056942914001\\
108.9997	1.14059809173604\\
109.0997	1.14062198661955\\
109.1997	1.14063462447133\\
109.2997	1.14065390797498\\
109.3997	1.14064335708696\\
109.4997	1.14066550831912\\
109.5997	1.14071350513399\\
109.6997	1.1407287104664\\
109.7997	1.14074913862543\\
109.8997	1.14074157455658\\
109.9996	1.14074390544773\\
110.0996	1.14077311163171\\
110.1996	1.14079701726136\\
110.2996	1.14084097136305\\
110.3996	1.14091748317375\\
110.4996	1.14105087545793\\
110.5996	1.14102402212052\\
110.6996	1.14106405412355\\
110.7997	1.14183060820583\\
110.8997	1.14296689217332\\
110.9997	1.14377157521213\\
111.0997	1.13642812423969\\
111.1997	1.10407365970695\\
111.2997	1.05110443545902\\
111.3997	0.982281725404427\\
111.4997	0.901906872162306\\
111.5997	0.803170142310618\\
111.6997	0.691052912986905\\
111.7997	0.570350094664094\\
111.8997	0.45225527377077\\
111.9997	0.342710340064204\\
112.0997	0.244882837781081\\
112.1997	0.161433829122911\\
112.2997	0.0986329371470531\\
112.3997	0.0532388640682253\\
112.4997	0.0202206239072178\\
112.5997	-0.00261835603743976\\
112.6997	-0.0178890090615875\\
112.7997	-0.0258031406327432\\
112.8997	-0.0277105716495132\\
112.9997	-0.029982557670251\\
113.0997	-0.0314376592666128\\
113.1997	-0.032814536810398\\
113.2997	-0.0331839718605429\\
113.3997	-0.0332553934294288\\
113.4997	-0.0332428812381931\\
113.5997	-0.0332143798702602\\
113.6997	-0.0331817085186179\\
113.7998	-0.0331621757468452\\
113.8998	-0.0331734351712869\\
113.9998	-0.0331678087787608\\
114.0998	-0.0331866533797014\\
114.1998	-0.0332065732474729\\
114.2998	-0.0332045775291003\\
114.3998	-0.0331835369372362\\
114.4998	-0.0331776040511627\\
114.5998	-0.0331658102369512\\
114.6998	-0.0331485319462563\\
114.7998	-0.0331553781901532\\
114.8998	-0.033163266799886\\
114.9997	-0.0331570210278016\\
115.0997	-0.0331286249001792\\
115.1997	-0.0331034685483589\\
115.2997	-0.0330713633221813\\
115.3997	-0.0330611655869666\\
115.4997	-0.0330591029108316\\
115.5997	-0.0330516321633916\\
115.6997	-0.0330517389396585\\
115.7998	-0.0330443129849174\\
115.8997	-0.0330531737962854\\
115.9997	-0.0330512500393656\\
116.0998	-0.0330449174913825\\
116.1998	-0.0330949258866255\\
116.2998	-0.0331045086706836\\
116.3998	-0.0331195238560802\\
116.4998	-0.0331342084551289\\
116.5998	-0.0331006948002559\\
116.6998	-0.0330723560650059\\
116.7998	-0.0330606566349108\\
116.8998	-0.033067346047583\\
116.9998	-0.0330833930569687\\
117.0998	-0.0330779693529084\\
117.1998	-0.0330579819557251\\
117.2998	-0.03303419665588\\
117.3998	-0.0330173119596398\\
117.4998	-0.033035069507906\\
117.5998	-0.0330677983302337\\
117.6998	-0.0330700389735499\\
117.7998	-0.0330670214416415\\
117.8998	-0.033041310071696\\
117.9998	-0.0330424105277094\\
118.0998	-0.0330274339666008\\
118.1998	-0.0330202381349547\\
118.2998	-0.0330200273111385\\
118.3998	-0.0329890168981595\\
118.4998	-0.0329827387299748\\
118.5998	-0.0329702844769457\\
118.6998	-0.0329644699522091\\
118.7998	-0.0329453933616888\\
118.8998	-0.0329454703826626\\
118.9998	-0.0329565670269611\\
119.0999	-0.0329750957414585\\
119.1999	-0.0329845552744248\\
119.2999	-0.0330183639603376\\
119.3999	-0.0330362687156174\\
119.4998	-0.0330414491358848\\
119.5999	-0.0330655312287929\\
119.6999	-0.0330759522099182\\
119.7999	-0.0330838988504951\\
};
\addplot [color=black,dashed,forget plot]
  table[row sep=crcr]{%
41.6291	-1\\
41.6291	7\\
};
\addplot [color=black,dashed,forget plot]
  table[row sep=crcr]{%
45	-1\\
45	7\\
};
\addplot [color=black,dashed,forget plot]
  table[row sep=crcr]{%
60	-1\\
60	7\\
};
\addplot [color=black,dashed,forget plot]
  table[row sep=crcr]{%
73.3591	-1\\
73.3591	7\\
};
\node[right, align=left, text=black]
at (axis cs:41.829,0.75) {\footnotesize Position fix};
\node[right, align=left, text=black]
at (axis cs:45.2,2.5) {\footnotesize Bag};
\node[right, align=left, text=black]
at (axis cs:60.2,2.5) {\footnotesize Pocket};
\node[right, align=left, text=black]
at (axis cs:73.559,0.75) {\footnotesize Position fix};
\end{axis}
\end{tikzpicture}%